\documentclass[10pt,twocolumn]{article}
\usepackage[dvips]{graphicx,graphics,color}
\usepackage{times,amsmath,amsfonts}

\newcommand{\Section}[1]{\vspace{-8pt}\section{\hskip -1em.~~#1}\vspace{-3pt}}

\newcommand{\bqn}{\begin{eqnarray}}
\newcommand{\eqn}{\end{eqnarray}}

\newcommand{\der}[2]{\frac{\partial #2}{\partial #1}}
\begin{document}

\date{October 31, 2002}

\title{\Large\bf Tensor-based Brain Surface Modeling and Analysis\footnote{Published in The proceedings of IEEE Conference on Computer
Vision and Pattern Recognition (CVPR) 2003 Vol I 467-473.\\ Note: Missing URL links and email addresses in the originally published version have been corrected.}}


\author{Moo K. Chung$^1$, Keith J. Worsley$^2$, Steve Robbins$^2$ and Alan C. Evans$^2$\\
$^1$Department of Statistics, Department of Biostatistics and Medical Informatics\\
Keck Laboratory for Functional Brain Imaging and Behavior\\
University of Wisconsin-Madison\\
Madison, WI 53706. USA\\
$^2$Montreal Neurological Institute, McGill University, Canada\\
\tt{mkchung@wisc.edu}}

\maketitle

\section*{\centering Abstract}

{\em We present a unified computational approach to tensor-based
morphometry in detecting the brain surface shape difference
between two clinical groups based on magnetic resonance images.
Our approach is novel in a sense that we combined surface
modeling, surface data smoothing and statistical analysis in a
coherent unified mathematical framework. The cerebral cortex has
the topology of a 2D highly convoluted sheet. Between two
different clinical groups, the local surface area and curvature of
the cortex may differ. It is highly likely that such surface shape
differences are not uniform over the whole cortex. By computing
how such surface metrics differ, the regions of the most rapid
structural differences can be localized. To increase the signal to
noise ratio, diffusion smoothing based on the explicit estimation
of Laplace-Beltrami operator has been developed and applied to the
surface metrics. As an illustration, we demonstrate how this new
tensor-based surface morphometry can be applied in localizing the
cortical regions of the gray matter tissue growth and loss in the
brain images longitudinally collected in the group of children.}

\Section{Introduction}

The cerebral cortex has the topology of a 2-dimensional convoluted
sheet. Most of the features that distinguish these cortical
regions can only be measured relative to that local orientation of
the cortical surface \cite{Dale}.  It is likely that different
clinical population will show different brain surface shape
differences \cite{Chung:02a, Chung:03a, MacDonald:00, Thompson:01, Thompson:00}. By computing how surface
metrics such as the cortical thickness, curvature and local
surface area differ among different groups, brain shape
differences can be quantified locally.

The first obstacle in developing surface-based morphometry is the
automatic segmentation of the cortical surfaces from magnetic
resonance images (MRI). It requires first correcting intensity
nonuniformity or RF inhomogeneity artifacts. We have used
nonparametric nonuniform intensity normalization method (N3),
which eliminates the dependence of the field estimate on anatomy
\cite{Sled}. The next step is the tissue classification into three
types:
 gray matter, white matter
 and cerebrospinal fluid (CSF). An artificial neural network classifier
 \cite{Ozkan, Vasken} or a mixture model cluster analysis
 \cite{Good} can be used to segment the tissue types automatically.
 After the tissue classification, the
  cortical surface is usually generated as a smooth triangular mesh. The most widely used
  method for triangulating the surface is the marching cubes algorithm
  \cite{Lorensen}. Level set method \cite{Sethian} or deformable surfaces
  method \cite{Davatzikos}  are also available. In our study, we have used
the anatomic segmentation using proximities (ASP) method
\cite{MacDonald:00}, which is a variant of the deformable surfaces
method, to generate cortical triangular meshes that has the
topology of a sphere. Brain substructures such as the brain stem
and the cerebellum were removed. Then an ellipsoidal mesh that
already had the topology of a sphere was deformed to fit the
shape of the cortex guaranteeing the same topology. The resulting
triangular mesh will consist of 40,962 vertices and 81,920
triangles with the average internodal distance of $3 \mbox{ mm}$.
Partial voluming is a problem with the tissue classifier but
topology constraints used in ASP method were shown to provide
some correction by incorporating many neuroanatomical {\em a
priori} information \cite{MacDonald:00}. The
triangular meshes are not constrained to lie on voxel boundaries.
Instead the triangular meshes can cut through a voxel, which can
be considered as correcting where the true boundary ought to be.
Once we have a triangular mesh as the realization of the cortical
surface, we can model how the cortical surface deforms over time.

In modeling the surface deformation, that is required in
comparing two different brain images, a proper mathematical
framework might be found in both differential geometry and fluid
dynamics. The concept of the {\em evolution of phase-boundary}, which describes the
geometric properties of the evolution of boundary layer between
two different materials due to internal growth or external force,
can be used to derive the mathematical formulation on the surface
deformation \cite{Chung:02a}. It is natural to assume the cortical surfaces to be a
smooth 2-dimensional {\em Riemannian manifold} parameterized by
two parameters \cite{Dale, Davatzikos}. Surface
parameterization of the cortical surface has been done previously
by \cite{Joshi}. From the surface parameterization, Gaussian and
mean curvatures of the brain surface can be computed and used to
characterize its shape \cite{Dale, Griffin, Joshi}. In
particular, \cite{Joshi} used the quadratic surface in estimating
the Gaussian and mean curvature of the cortical surfaces. Surface
parameterization enables us to compute surface metrics such as
local area dilatation and curvature differences that characterize
the surface shapes.

Due to errors in image intensities, surface extraction and
surface fitting, surface-based signal smoothing is required to
increase the signal-to-noise ratio. We have used diffusion
smoothing or the Laplace-Beltrami flow to smooth surface metrics.
Although diffusion smoothing has been used widely in image
analysis \cite{Malladi, Perona, Sochen, Tang, Taubin}, there is only two papers
so far that use the Laplace-Beltrami flow to smooth out brain surface
data \cite{Andrade, Chung:03a}. Based on the finite element method (FEM), we
explicitly estimate the Laplace-Beltrami operator and then the
finite difference scheme is used to iteratively solve a diffusion
equation on the surface. Because the Laplace-Beltrami operator is
estimated as a linear weight of the neighboring function values,
once the linear weights are computed at the beginning, it will be
repeatedly used in the subsequent iterations; hence, avoiding
sparse matrix inversions which are required in most FEM
formulation. For surface-based statistical inference, the
smoothing filter size has been incorporated into the $P$-value computation
based on random fields theory \cite{Worsley:96a}.

As an illustration of our unified approach to tensor-based surface
morphometry, we will demonstrate how the surface-based
statistical analysis can be applied in localizing the cortical
regions of tissue growth and loss in brain images longitudinally
collected in a group of children and adolescents.

\section{Surface Medeling}
Let ${\bf U}^i({\bf x})=(U_1^i,U_2^i,U_3^i)^t$ be the 3D
displacement vector required to deform the structure at ${\bf
x}=(x_1,x_2,x_3)$ in the gray matter of the template brain
$\Omega_{atlas}$ to the homologous structure in subject image
$\Omega^i$. We assume that the whole gray matter volume in
$\Omega_{atlas}$ will deform continuously and smoothly to
$\Omega^i$ via the deformation ${\bf x} \to {\bf x}+{\bf U}^i$
while the cortical boundary $\partial \Omega_{atlas}$ will deform
to $\partial \Omega^i$. The cortical surface $\partial \Omega^i$
may be considered as consisting of two parts: the outer cortical
surface $\partial \Omega_{out}^i$ between the gray matter and CSF
and the inner cortical surface $\partial \Omega_{in}^i$ between
the gray and white matter, i.e.
$$\partial \Omega^i=\partial \Omega_{out}^i \cup \partial \Omega_{in}^i.$$
We propose the following stochastic model on the displacement
${\bf U}^i$: \bqn {\bf U}^i({\bf x}) = \mu({\bf x}) +
\Sigma^{1/2}({\bf x})\epsilon({\bf x}), \; {\bf x} \in
\partial \Omega_{atlas}, \label{eq:model}\eqn where ${\bf \mu}$ is the mean
displacement and $\Sigma^{1/2}$ is the covariance matrix, which
allows for correlations between components of the displacement
fields. The components of the error vector $\epsilon$ are are
assumed to be independent and identically distributed as smooth
stationary Gaussian random fields with zero mean and unit
variance.

\begin{figure}
\includegraphics[scale=0.315]{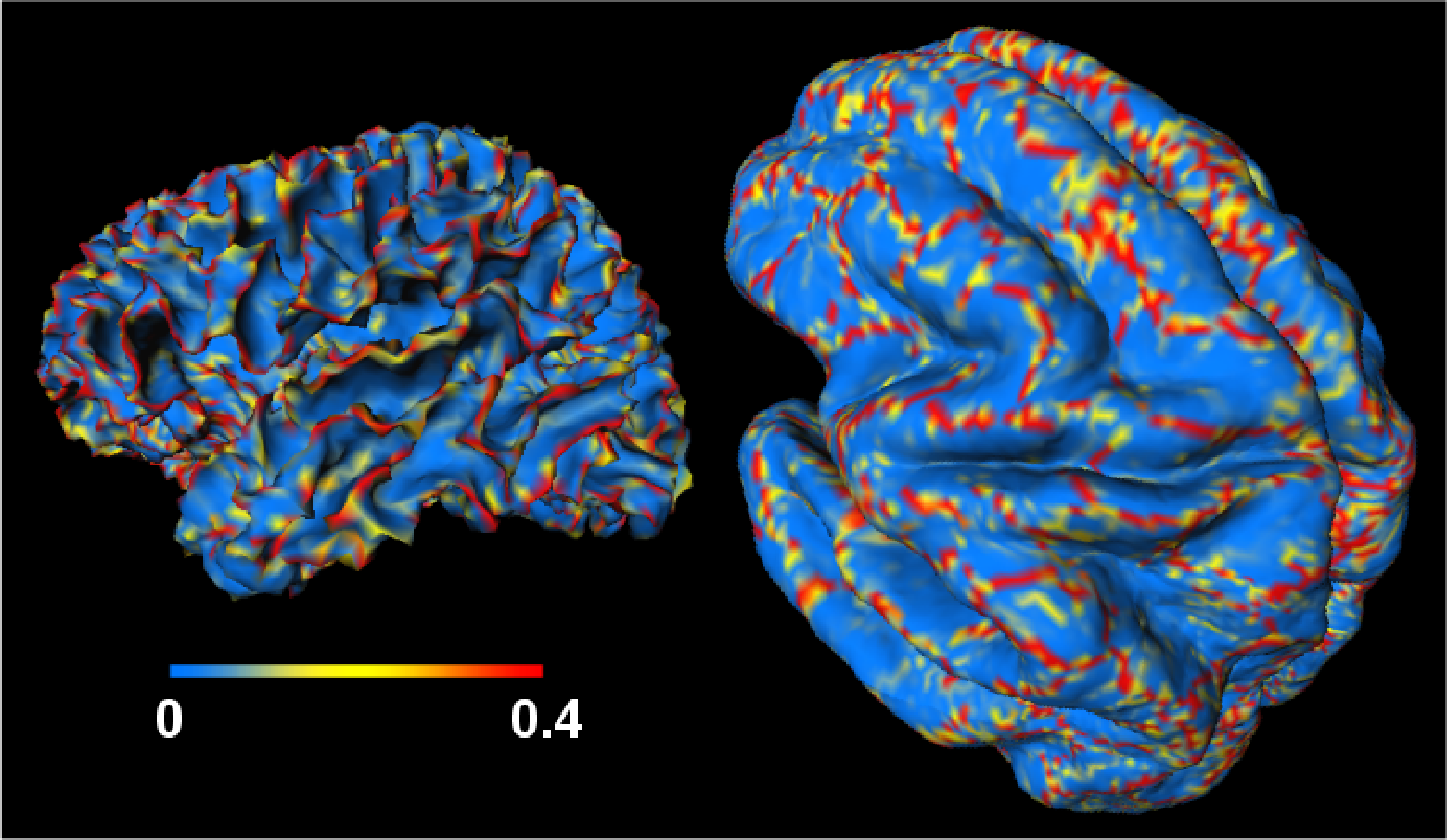}
\caption{\label{fig:gyrus} Individual gyral patterns mapped onto
the template surface $\partial \Omega_{atlas}$. The gyri of the
subject match the gyri of the template illustrating a close homology
between the surface of an individual subject and the template
surface. The Gyri are extracted by thresholding the thin-plate
spline energy functional on the inner surface. If there is no
homology between the corresponding vertices, we would have
complete misalignment.}
\end{figure}

Estimating the surface displacement fields ${\bf U}^{ij}: \partial
\Omega^i \to \partial \Omega^j$ between two images $i$ and $j$,
and the surface extraction can be performed at the same time.
This method works best in the case of matching two images of a
single subject taken at different times. First, an ellipsoidal
mesh placed outside the brain was shrunk down to the surface
$\partial \Omega_{in}^i$. The vertices of the resulting inner
mesh are indexed and the ASP algorithm will deform the inner mesh
to fit the outer surface $\partial \Omega_{out}^i$ by minimizing
a cost function that involves bending, stretch and other
topological constrains \cite{MacDonald:00}. The vertices indexed
identically on both meshes will lie within a very close proximity
and these define the automatic linkage in the ASP algorithm. To
generate the outer surface $\partial \Omega_{out}^j$, we start
with the inner surface $\partial \Omega_{in}^i$, and then deform
it to match the outer surface $\partial \Omega_{out}^j$ by
minimizing the same cost function. Starting with the same mesh in
two outer surface extractions, each point on $\partial
\Omega_{in}^i$ gets mapped to corresponding points on $\partial
\Omega_{out}^i$ and $\partial \Omega_{out}^j$ giving us the outer
surface deformation ${\bf U}^{ij}: \partial \Omega_{out}^i \to
\partial \Omega_{out}^j$. We do not use inner surface deformation in
our study although estimating inner surface deformation can be done
similarly. This method assumes that the shape of the cortical
surface does not appreciably change between images $\Omega^i$ and $\Omega^j$.
This assumption is valid in the case of brain development for a
short period of time, where it can be shown that the
within-subject deformation field is substantially smaller than
between-subject deformation.

Constructing surface template $\partial \Omega_{atlas}$, where
statistical parametric maps (SPM) of surface metrics will be
formed, is done by averaging the coordinates of corresponding
vertices that have the same indices. This surface atlas
construction method has been first introduced by
\cite{MacDonald:00}, where it is used to create the cortical
thickness map for 150 normal subjects. The geometrical
constraints such as stretch and bending terms in ASP algorithm
enforces a relatively consistent correspondence on the cortical
surface. Figure \ref{fig:gyrus} shows the mapping of gyral
pattern (red and yellow lines) of a single subject onto the atlas
surface. The gyri of the subject matches the gyri of the atlas.
Note the full anatomical details still presented in $\partial
\Omega_{atlas}$ even after the vertex averaging. Major sulci such
as the central sulcus and superior temporal sulcus are clearly
identifiable. If there is no homology between corresponding
vertices, one would only expect to see featureless dispersion of
points.

Once we have extracted triangular surface meshes and established
mapping from a vertex in $\partial \Omega^i$ to a corresponding
vertex in $\partial \Omega^j$, the next step is modeling and
computing the metric tensor differences between two surfaces. In
order to compute metric tensors, {\em surface parameterization} is
needed. We model the cortical surface as a smooth 2D Riemannian
manifold parameterized by two parameters $u^1$ and $u^2$ such that
any point ${\bf x} \in
\partial \Omega^i$ can be uniquely represented as
${\bf x} = {\bf X}({\bf u})$ for some parameter space ${\bf
u}=(u^1,u^2) \in D \subset \mathbb{R}^2$. A quadratic polynomial
\bqn z = \beta_1u^1 + \beta_2u^2 + \beta_3(u^1)^2 + \beta_4u^1u^2
+ \beta_5(u^2)^2 \label{eq:quad_poly}\eqn was used as a local
parameterization fitted via the least-squares estimation on the
tangent plane. Using the least-squares method, these coefficients
$\beta_i$ can be estimated. Slightly different quadratic surface
parameterizations are used in estimating
 curvatures of a macaque monkey brain surface \cite{Joshi} \cite{Khaneja}.
 Once $\beta_i$ are estimated,
\bqn {\bf X}(u^1,u^2) &=& \big(u^1,u^2,z(u^1,u^2)\big)^t
\label{eq:para}\eqn
   becomes a local surface
  parameterizations of choice. Even thought metric tensors
  depend on the choice of parameterization, local surface area and
  curvature dilatation, which are introduced in the next
  section, are independent of parameterization.

\Section{Metric Tensor Computation} We introduce the concepts of
surface area and curvature dilatation, which can be used in
quantifying surface shape differences. Suppose that ${\bf x} =
{\bf X}({\bf u})$ is the parameterization of surface $\partial
\Omega$. Let ${\bf X}_i=\partial {\bf X}\slash
\partial u^i$. From (\ref{eq:para}), ${\bf X}_i$ are given in terms
of coefficients $\beta_i$ and the {\em Riemannian metric tensor}
$g_{ij}$ is given by the inner product between two vectors ${\bf
X}_i$ and ${\bf X}_j$, i.e. $g_{ij} = \langle {\bf X}_i, {\bf X}_j
\rangle$. The Riemannian metric tensor $g_{ij}$ measures the
amount of the deviation of the cortical surface from a flat
Euclidean plane. The Riemannian metric tensor enables us to
quantify lengths, angles and areas in the cortical surface. Let $g
=(g_{ij})$ be a $2 \times 2$ metric tensor matrix. Then the total
surface area of the cortex $\partial \Omega$ is given by
$$\|\partial \Omega\| = \int_D
\sqrt{\det g} \;d{\bf u},$$ where $D=X^{-1}(\partial \Omega)$ is
the parameter space \cite{Kreyszig}. The integrand $\sqrt{\det g}$
is called the {\em infinitesimal surface area element} and it
measures the area of the unit square in the parameter space $D$,
that has been transformed via $X: D \to
\partial \Omega$. The infinitesimal surface area element is
a generalization of Jacobian. The {\em local surface area
dilatation} $\Lambda_{area}$ from $\partial \Omega^i$ to $\partial
\Omega^j$, whose metric tensor matrices are given by $g_i$ and
$g_j$, is then defined as \bqn \Lambda_{area} = \frac{\sqrt{\det
g_j}-\sqrt{\det g_i}}{\sqrt{\det g_i}},\label{eq:area-element}\eqn
which measures percentage local area differences. The dilatation
is invariant under parameterization, i.e. the area dilatation
 is the same no matter which parameterization is chosen.

Instead of using metric tensors $g_{ij}$, it is possible to
formulate local surface area change in terms of the areas of the
corresponding triangles. However, this formulation assign surface
area change values to each face instead of each vertex and this
causes problems in both surface-based smoothing and statistical
analysis, where values are defined on vertices. Defining scalar
values on vertices from face values can be done by the weighted
average of face values, which should converge to
(\ref{eq:area-element}). It is not hard to develop surface-based
smoothing and statistical analysis on values defined on faces but
traditionally surface metrics are computed on vertices.

\begin{figure}
\includegraphics[scale=0.485]{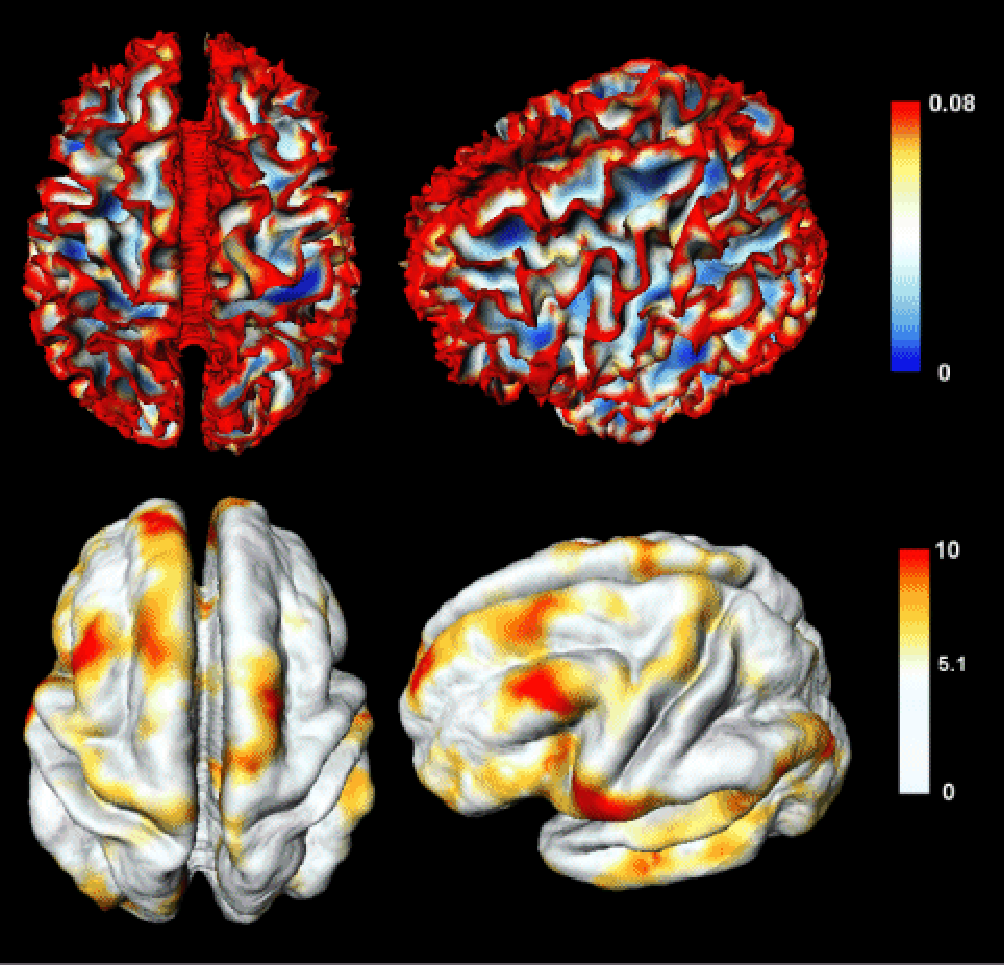}
\caption{\label{fig:bending} Top: The thin-plate spline energy functional
computed on the inner surface of a 14 year old subject. It
measures the amount of folding in the cortical surface. Bottom:
$t$ statistical map showing statistically significant region of
curvature increase ($t > 5.1$) over time between ages 12 and 16.
Most of curvature increase occurs on gyri while there is no
significant change of curvature on most of sulci.}
\end{figure}

Curvatures of the surface can be also used to quantify the surface
shape difference. The {\em principal curvatures} can characterize
the shape and location of the sulci and gyri, which are the
valleys and crests of the cortical surfaces \cite{Bartesaghi, Joshi, Khaneja, Subsol:99}. By measuring the
curvature changes, rapidly folding and cortical regions can be
localized. Let $\kappa_1$ and $\kappa_2$ be the two principal
curvatures as defined in \cite{Kreyszig}. The principal curvatures
can be represented as functions of $\beta_i$s in quadratic surface
(\ref{eq:quad_poly}) \cite{Kreyszig}. To measure the amount of
folding, we define curvature metric $K$ as a function of the
principal curvatures: $K= (\kappa_1^2 + \kappa_2^2)/2 + \alpha$.
We may arbitrarily set $\alpha =0.001$. $\alpha$ is added to make
sure that the curvature dilatation is well defined. The mean of
the square of the principal curvatures is usually refereed as the
thin-plate spline energy functional. If the cortical surface is
flat, curvature metric $K$ obtains the minimum 0.001. The larger
the curvature metric, the more surface will be crested as shown in
Figure \ref{fig:bending}. The {\em local curvature dilatation
rate} $\Lambda_{curvature}$ is similarly defined as
(\ref{eq:area-element}).
\Section{Diffusion Smoothing} In order to increase the
signal-to-noise ratio (SNR) and to satisfy Gaussian random field assumptions that is required in
our statistical analysis \cite{Worsley:96a}, surface-based signal smoothing or
filtering is needed. One important reason that the Gaussian kernel
smoothing is widely used in brain imaging analysis is that it
preserves the Gaussian noise assumption even after the filtering.
So if we are assuming a linear model with a Gaussian error, the
Gaussian kernel smoothed image will still follow the same linear
model but with a more smooth and isotropic covariance structure.
By smoothing the data on the cortical surface, the SNR will
increase and in turn it will be easier to localize the
morphological changes. However, due to the convoluted nature of
the cortex whose geometry is non-Euclidean, it is not possible to
apply Gaussian kernel smoothing on the cortical surface directly.
Gaussian kernel smoothing of functional data $f({\bf x}), {\bf
x}=(x_1,\dots,x_n) \in {\mathbb{R}}^n$ with FWHM ({\em full width
at half maximum}) $=4(\ln 2)^{1/2}\sqrt t$ is defined as the
convolution of the Gaussian kernel with $f$: \bqn F({\bf x},t)=
\frac{1}{(4\pi t)^{n/2}}\int_{{\mathbb{R}}^n} e^{-(x-y)^2/4t} f(y)
\; dy. \label{eq:convolution}\eqn  Since formulation
(\ref{eq:convolution}) can not be directly applied to the cortical
surfaces, we reformulate it as a solution of a diffusion equation
on a Riemannian manifold. This generalization is called {\em
diffusion smoothing} and has been used in the analysis of fMRI
data on the cortical surface \cite{Andrade}. In many computer
vision areas, it is usually refereed as simply {\em diffusion} or {\em Beltrami flow}
\cite{Malladi, Sochen}. It can be shown that (\ref{eq:convolution}) is the
integral solution of an isotropic diffusion equation $
\partial_t F = \Delta F$ with the initial condition
$F({\bf x},0)=f({\bf x})$, where $\Delta$ is the $n$-dimensional
Euclidean Laplacian. Generalizing the Euclidean Laplacian to an
arbitrary Riemannian manifold, we get the {\em Laplace-Beltrami
operator} \cite{Kreyszig}. The approach taken in \cite{Andrade} is
based on a local flattening of the cortical surface and estimating
the planar Laplacian, which may not be as accurate as our
estimation based on the finite element method (FEM). Further, our
explicit FEM approach completely avoid any local or global surface
flattening. For given Riemannian metric tensor $g_{ij}$, the
Laplace-Beltrami operator $\Delta$ is given as \bqn \Delta F =
\sum_{i,j}
\frac{1}{|g|^{1/2}}\der{u^i}{}\Big(|g|^{1/2}g^{ij}\der{u^j}{F}\Big)
\label{eq:Laplace-Beltrami}, \eqn where $(g^{ij})=g^{-1}$
\cite{Kreyszig}. Using the FEM on
the triangular cortical mesh generated by the ASP algorithm, it is
possible to estimate the Laplace-Beltrami operator as the linear
weights of neighboring vertices \cite{Chung:02a}.

\begin{figure}
\centering
\includegraphics[scale=0.3]{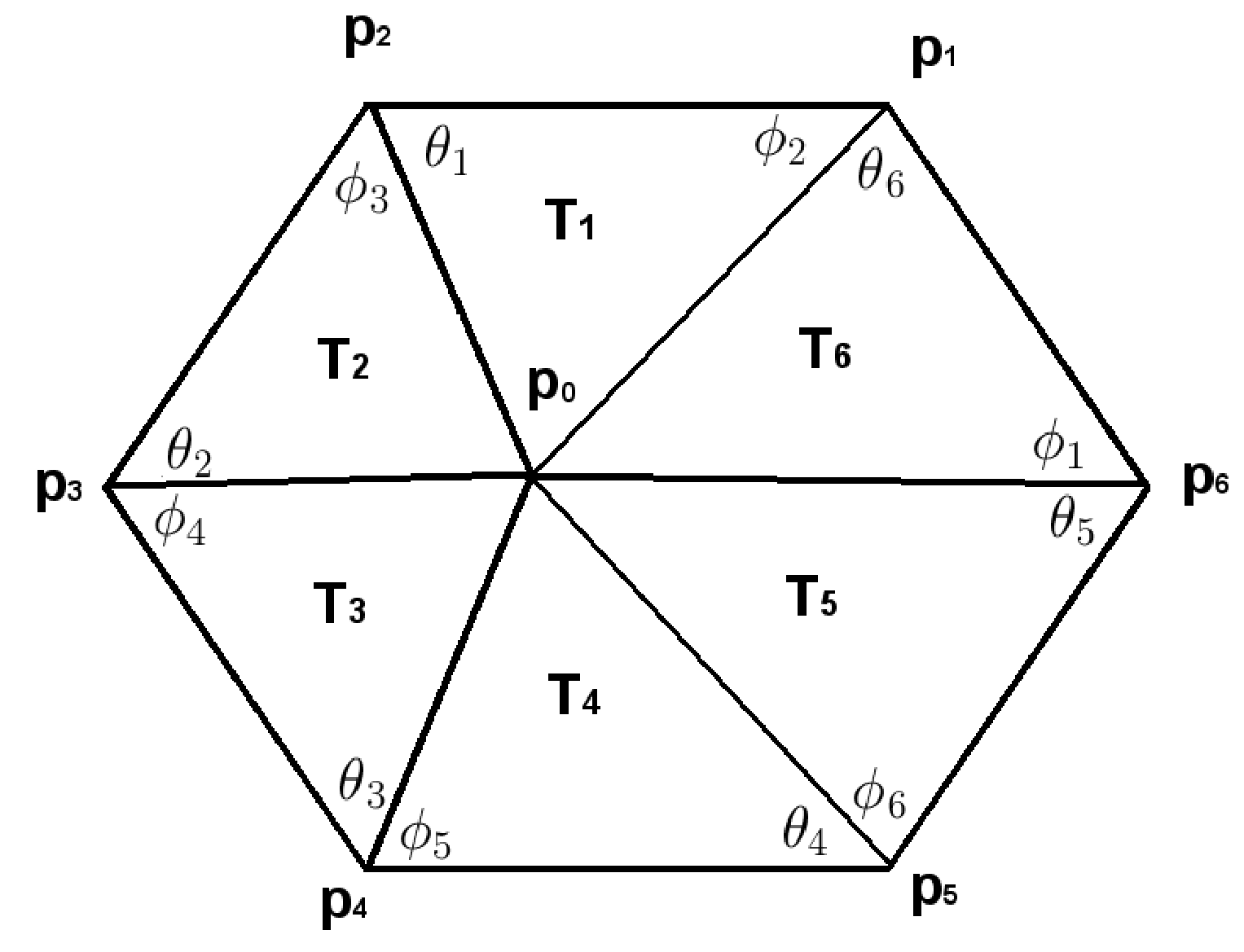}
\caption{\label{fig:hexagon1} A typical triangulation in the
neighborhood of ${\bf p} ={\bf p}_0$. When ASP algorithm is used,
 the triangular mesh is constructed in such a way that it is always pentagonal
 or hexagonal.}
\end{figure}
\begin{figure}
\includegraphics[scale=0.4]{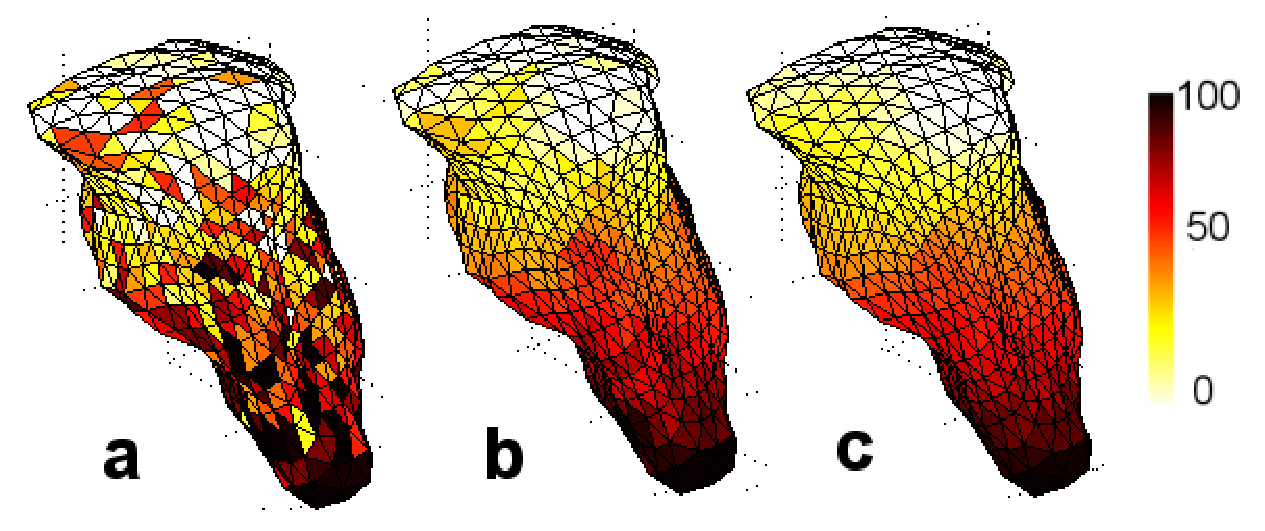}
\caption{\label{fig:smoothing}Diffusion smoothing simulation on a
triangular mesh consisting of 1280 triangles. This smaller mesh is
the surface of the brain stem. The artificial signal was generated
with Gaussian noise to illustrate the smoothing process. (a) The
initial signal. (b) After 10 iterations with $\delta t= 0.5$. (c)
After 20 iterations with $\delta t= 0.5$.}
\end{figure}

Let ${\bf p}_1,\cdots,{\bf p}_m$ be $m$ neighboring vertices
around the central vertex ${\bf p}={\bf p}_0$. Then the explicit
linear estimation of the Laplace-Beltrami operator based on FEM is
given by
$$\widehat {\Delta F}({\bf p}) = \sum_{i=1}^m
w_i\big(F({\bf p}_i)-F({\bf p})\big)$$ with the weights
$$w_i= [\cot \theta_i + \cot \phi_i]/\sum_{i=1}^m \|T_i\|,$$
where  $\theta_i$ and $\phi_i$ are the two angles opposite to the
edge connecting ${\bf p}_i$ and ${\bf p}$, and $\|T_i\|$ is the
area of the $i$-th triangle (Figure \ref{fig:hexagon1}). This is
an improved formulation from the previous study \cite{Andrade}
that uses diffusion smoothing on the cortical surface, where the
Laplacian is simply estimated as the planar Laplacian after
locally flattening the triangular mesh consisting of nodes ${\bf
p}_0,\cdots,{\bf p}_m$ onto a flat plane. In the numerical
implementation, we have used formula
$$2\cot \theta_i = \langle {\bf p}_{i+1}-{\bf p}, {\bf p}_{i+1} - {\bf p}_i\rangle /\|T_i\|,$$
$$2\cot \phi_i = \langle {\bf p}_{i-1}-{\bf p} , {\bf
p}_{i-1} - {\bf p}_i\rangle/\|T_i\|$$ and $\|T_i\| = \|({\bf
p}_{i+1} - {\bf p}) \times ({\bf p}_i - {\bf p}) \|/2.$
Afterwards, the finite difference scheme is used to iteratively
solve the diffusion equation at each vertex ${\bf p}$:
$$F({\bf p},t_{n+1})= F({\bf p},t_n) + (t_{n+1}-t_n)
\widehat{\Delta} F({\bf p},t_n),$$ with the initial condition
$F({\bf p},t_0)=f({\bf p})$. $N$-iterations are equivalent to the
diffusion of the initial data $f$ for duration
$N\delta t$. If the diffusion were applied to Euclidean space, it
would be equivalent to Gaussian kernel smoothing with $\mbox{FWHM}
=4(\ln2)^{1/2}\sqrt{N\delta t}.$ The iteration step size $\delta
t$ is chosen to satisfy $ \delta t \leq \min
(A,B)/\widehat{\Delta} F(p,t_n)$ for all $n$, where $A= | \max_i F(p_i,t_n)-F(p,t_n)|$ and $B= | \min_i F(p_i,t_n)-F(p,t_n)|$, to
guarantee the convergence. Computing the linear weights for the
Laplace-Beltrami operator takes a fair amount of time (four
minutes in $\tt{MATLAB}$ running on a Pentium III machine), but once the
weights are computed, it is applied through the whole iteration
repeatedly and the actual finite difference scheme takes only two
minutes for 100 iterations. Figure \ref{fig:smoothing} illustrates
the process of diffusion smoothing on the surface of the
substructure of the brain (brain stem).

\Section{Brain Surface Data Analysis} Under the assumption of
stochastic model (\ref{eq:model}), it can be shown that the area
dilatation is approximately distributed as Gaussian: \bqn
\Lambda({\bf x}) = \lambda ({\bf x}) + \epsilon({\bf x}),
\label{eq:model2}\eqn
 where $\lambda =\text{tr }[g^{-1}(\nabla X)^t (\nabla \mu) \nabla X]$
 is the mean area dilatation and error $\epsilon$ is
 a mean zero Gaussian random field defined on the cortical surface. The curvature dilatation can be
 modeled similarly. These theoretical model assumptions have been
 verified using Lilliefors test at $0.05$ level \cite{Chung:03a}.
 From statistical model (\ref{eq:model2}), we are interested in
 testing the hypothesis: $H_0: \lambda({\bf x})=0 \text{ for all } {\bf x} \in \partial
\Omega_{atlas}$, v.s. $H_1: \lambda({\bf x}) \neq 0 \text{ for some } {\bf x} \in \partial \Omega_{atlas}.$ The maximum of $T$ random field will be used as a test
statistic \cite{Worsley:96a}. The $T$ random field on manifold
$\partial \Omega_{atlas}$ is defined as
$$T({\bf x}) = \frac{M({\bf x})}{S({\bf x})/\sqrt{n}},\;
{\bf x} \in \partial \Omega_{atlas}$$ where $M$ and $S$ are the
sample mean and standard deviation of metric $\Lambda$. $T({\bf
x})$ is distributed as a student's $t$ with $n-1$ degrees of
freedom at each voxel ${\bf x}$. The $P$-value can be
approximated asymptotically \cite{Worsley:96a}. For high threshold
$y$, it can be shown that \bqn P\Big( \max_{{\bf x} \in
\partial\Omega_{atlas}} T({\bf x}) \geq y\Big) \approx
\sum_{i=0}^3 \phi_i(\partial \Omega_{atlas})\rho_i(y),\eqn where
$\rho_i$ is the $i$-dimensional EC-density and the Minkowski
functional $\phi_i$ for $\partial \Omega_{atlas}$ are
$$\phi_0=2, \;
\phi_1=0, \; \phi_2= \|\partial \Omega_{atlas} \|, \; \phi_3=0$$
and $\|\partial \Omega_{atlas}\|$ is the total surface area of
$\partial \Omega_{atlas}$ \cite{Worsley:96a}. When diffusion
smoothing with given FWHM is applied to metric $\Lambda$ on
surface $\partial \Omega_{atlas}$, the 0-dimensional and
2-dimensional EC-density becomes
$$\rho_0(y)=\int_y^{\infty}
\frac{\Gamma(\frac{n}{2})}{((n-1)\pi)^{1/2}\Gamma(\frac{n-1}{2})}
\Big(1+\frac{y^2}{n-1}\Big)^{-n/2}\; dy,$$
$$\rho_2(y)=\frac{1}{\text{FWHM}^2}\frac{4 \ln 2}{(2\pi)^{3/2}}
\frac{\Gamma(\frac{n}{2})y\Big(1+\frac{y^2}{n-1}\Big)^{-(n-2)/2}}{(\frac{n-1}{2})^{1/2}
\Gamma(\frac{n-1}{2})}.$$ Therefore, the $P$-value can be
approximated by
$$P\Big( \max_{{\bf x} \in \partial\Omega_{atlas}} T({\bf x})
\geq y\Big) \approx 2\rho_0(y) + \|\partial \Omega_{atlas}
\|\rho_2(y).$$ For one-sided $\alpha$-level test, we numerically
solve equation $2\rho_0(y) + \|\partial \Omega_{atlas} \|\rho_2(y)
=\alpha$ and reject $H_0$ if $T \geq y$ or $T \leq -y$.

\begin{figure}
\includegraphics[scale=0.445]{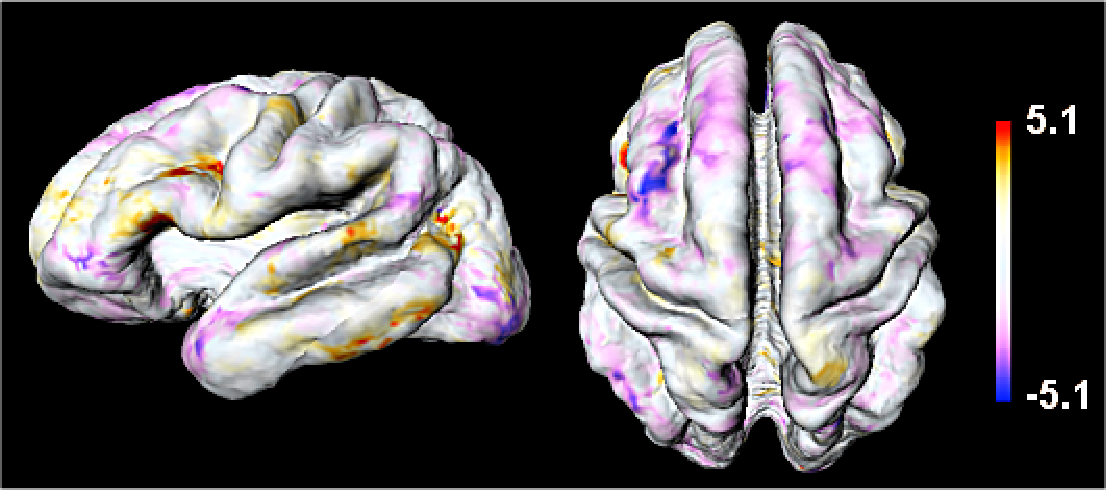}
\caption{\label{fig:corticalarea} $t$-map of the cortical surface
area dilatation showing the statistically significant region of
area expansion and reduction over time. The red regions are
statistically significant surface area expansions while the blue
regions are statistically significant surface area reductions
between ages 12 and 16.}
\end{figure}

\Section{Applications} Two T$_1$-weighted MR scans were acquired
for 28 normal subject at different times on the GE Sigma 1.5-T
superconducting magnet system. The first scan was obtained at the
age $t_1=11.5 \pm 3.1$ years and the second scan was obtained at
the age $t_2=16.1 \pm 3.2$ years. We are interested in detecting
the regions of the cortical shape difference over time. We compute
the total surface area $\|\partial \Omega_{atlas} \|$ by summing
the area of each triangle in a triangulated surface. The total
surface area of the average atlas brain is 275,800 mm$^2$, which
is roughly the area of $53 \times 53 \; \mbox{cm}^2$ sheet. We
also computed the local area and the curvature dilatations.
Surface metrics are then filtered with $20 \mbox{ mm}$ FWHM
diffusion smoothing. At $\alpha=0.025\%$ level, statistically
significant regions of local area and curvature difference over
time are detected. Figure \ref{fig:bending} shows the superior
frontal and middle frontal gyri curvature increase over time.
Figure \ref{fig:corticalarea} shows local surface expansion in
Broca's area in the left hemisphere and local surface shrinkage in
the left superior frontal sulcus. Most of surface reduction are
concentrated near the frontal region.  It is interesting to note
that between these two gyri we have detected local surface area
decrease. It might be possible that
 local surface area shrinking in the superior frontal sulcus
causes the bending in the neighboring middle and superior frontal
gyri. While the gray matter is shrinking in both total surface
area and volume, the cortex itself seems to get folded to give
increasing curvature in brain development for children.

To verify that our modeling and analysis do not detect any false
signal, our methods have been checked on null data. The null data
is created by reversing time for randomly chosen half of the
subjects. In the null data, the mean time difference $t_2 -t_1$ is
$-0.24$ year so the statistical analysis presented here should not
detect any morphological changes. In fact, we did not detect any
statistically significant morphological changes.
\Section{Conclusions} The unified tensor-based surface morphometry
presented here can localize the regions of surface shape
difference between two clinical groups of magnetic resonance
images at a local level without specifying the regions of interest
or landmarks. The approach avoids artificial surface flattening,
which may destroy the inherent geometrical structure of the
cortical surface. Our metric tensor formulation gives us an added
advantage that not only it can be used to measure local surface
area and curvature change of the cortex but also it is used for
generalizing Gaussian kernel smoothing on the cortex via diffusion
smoothing. Since it is a direct generalization of Gaussian kernel
smoothing, the diffusion smoothing should locally inherit many
mathematical and statistical properties of Gaussian kernel
smoothing applied to standard 3D whole brain volume. The novelty
of our diffusion smoothing is that we used the explicit estimation
of the Laplace-Beltrami operator. We succeeded in combining and
unify surface modeling, morphometry, image smoothing and
statistical inference in the same mathematical framework.
\section*{Acknowledgments}
Authors wish to thank Jothan Taylor of the department of statistics, Stanford University for many discussions on diffusion smoothing. Authors also wish to thank Thomas Paus of Montreal Neurological Institute, Montreal. Jay Giedd and Judith Raporport at National Institute of Mental Health had provided the original 28 subject brain images.

\end{document}